\documentclass[journal]{IEEEtran}
\usepackage{blindtext}
\usepackage{graphicx}
\usepackage{amssymb}
\usepackage{dsfont}
\usepackage{multirow}
\usepackage{amsmath}
\usepackage{float}
\usepackage{subfigure}
\usepackage[section]{placeins}
\usepackage{array}
\usepackage{colortbl}
\usepackage{color}
\newcolumntype{L}[1]{>{\raggedright\let\newline\\\arraybackslash\hspace{0pt}}m{#1}}
\newcolumntype{C}[1]{>{\centering\let\newline\\\arraybackslash\hspace{0pt}}m{#1}}
\newcolumntype{R}[1]{>{\raggedleft\let\newline\\\arraybackslash\hspace{0pt}}m{#1}}
\ifCLASSINFOpdf
\else
\fi
\usepackage{url}
\begin{document}
%
\title{Unsupervised Learning for Cell-level Visual Representation in Histopathology Images with Generative Adversarial Networks}

\author{Bo Hu$^\sharp$, Ye Tang$^\sharp$, Eric I-Chao Chang, Yubo Fan, Maode Lai and Yan Xu*

\thanks{This work is supported by the Technology and Innovation Commission of Shenzhen in China under Grant shenfagai2016-627, Microsoft Research under the eHealth program, the National Natural Science Foundation in China under Grant 81771910, the National Science and Technology Major Project of the Ministry of Science and Technology in China under Grant 2017YFC0110903, the Beijing Natural Science Foundation in China under Grant 4152033, Beijing Young Talent Project in China, the Fundamental Research Funds for the Central Universities of China under Grant SKLSDE-2017ZX-08 from the State Key Laboratory of Software Development Environment in Beihang University in China, the 111 Project in China under Grant B13003. \emph{* indicates corresponding author; $^\sharp$ indicates equal contribution.}}
\thanks{Bo Hu, Ye Tang, Yubo Fan and Yan Xu are with the State Key Laboratory of Software Development Environment and the Key Laboratory of Biomechanics and Mechanobiology of Ministry of Education and Research Institute of Beihang University in Shenzhen and Beijing Advanced Innovation Centre for Biomedical Engineering, Beihang University, Beijing 100191, China (email: bohu1996@gmail.com; yetang1995@gmail.com; yubofan@buaa.edu.cn; xuyan04@gmail.com).}
\thanks{Maode Lai is with the Department of Pathology, School of Medicine, Zhejiang
University (email: lmd@zju.edu.cn).}
\thanks{Eric I-Chao Chang, and Yan Xu are with Microsoft Research, Beijing 100080, China (email: echang@microsoft.com; v-yanx@microsoft.com).}
}
%
%
%

%
%

\markboth{}%
{Shell \MakeLowercase{\textit{et al.}}: Bare Demo of IEEEtran.cls for Journals}
%



\maketitle

\begin{abstract}

The visual attributes of cells, such as the nuclear morphology and chromatin openness, are critical for histopathology image analysis. By learning cell-level visual representation, we can obtain a rich mix of features that are highly reusable for various tasks, such as cell-level classification, nuclei segmentation, and cell counting. In this paper, we propose a unified generative adversarial networks architecture with a new formulation of loss to perform robust cell-level visual representation learning in an unsupervised setting. Our model is not only label-free and easily trained but also capable of cell-level unsupervised classification with interpretable visualization, which achieves promising results in the unsupervised classification of bone marrow cellular components. Based on the proposed cell-level visual representation learning, we further develop a pipeline that exploits the varieties of cellular elements to perform histopathology image classification, the advantages of which are demonstrated on bone marrow datasets.

\end{abstract}

\section{Introduction}
\begin{figure}[h]
\centering
\includegraphics[width=0.48\textwidth]{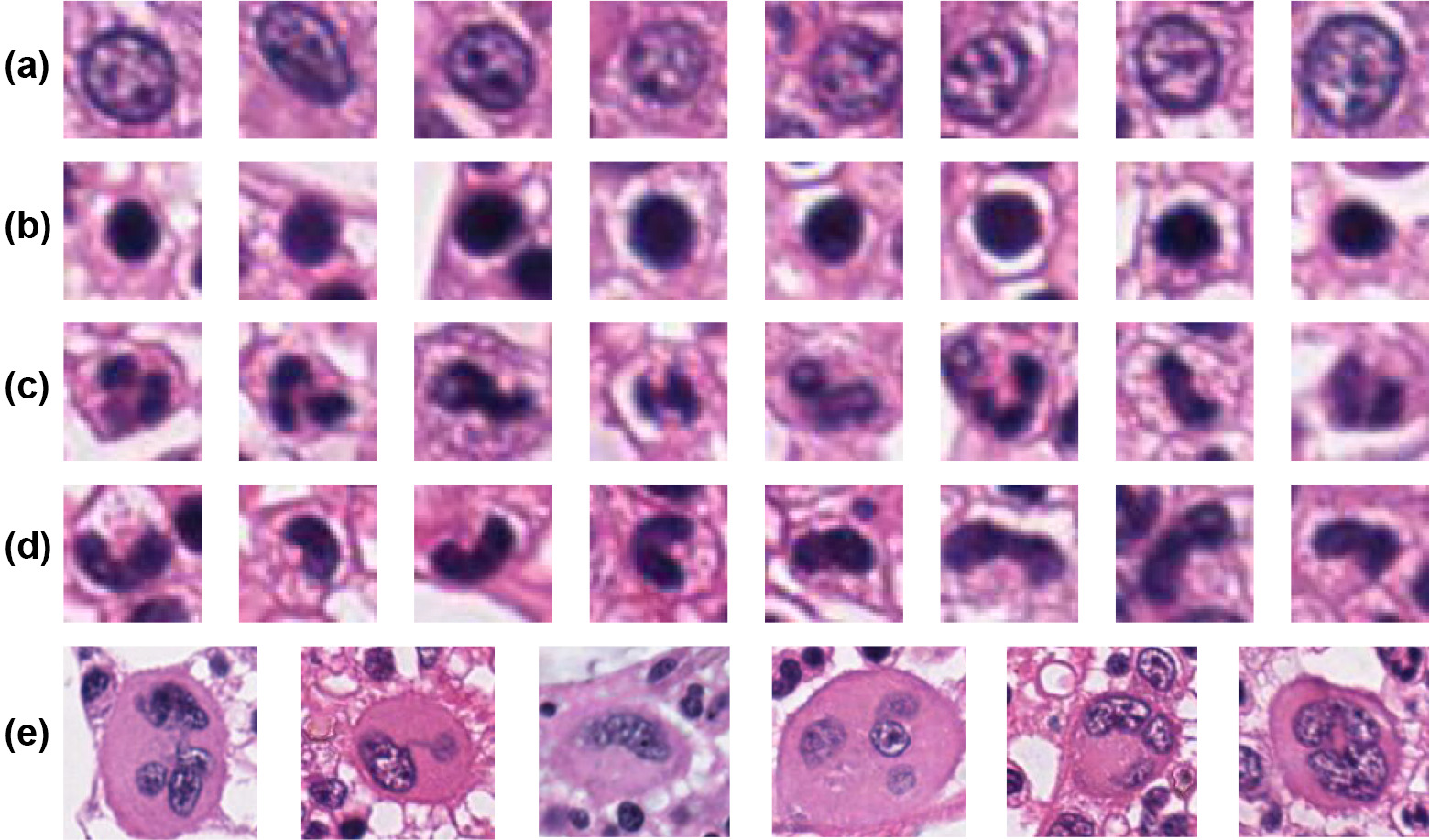}
\caption{Examples of five types of cellular elements in bone marrow: $\displaystyle (a)$ granulocytes precursors such as myeloblasts, $\displaystyle (b)$ cells with dark, dense, and close phased nuclei, the candidates of which are most likely lymphocytes and normoblasts, $\displaystyle (c)$ granulocytes such as neutrophils, $\displaystyle (d)$ monocytes, and $\displaystyle (e)$ megakaryocytes. Five types of cells can be distinguished by the chromatin openness, the density of nuclei, and if nuclei show the appearance of being segmented. Megakaryocytes appear the least often, as well are the most distinguished due to their massive size.}
\label{fig1}
\end{figure}

\begin{figure}[h]
\centering 
\subfigure[abnormal]{ \label{fig:a} 
\includegraphics[width=0.22\textwidth]{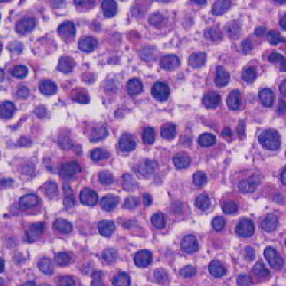} 
} 
\subfigure[normal]{ \label{fig:b} 
\includegraphics[width=0.22\textwidth]{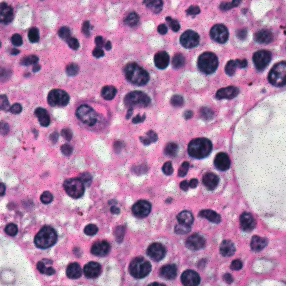} 
} 
\caption{Examples of bone marrow images sliced from Whole Slide Images (WSI). Too many myeloblasts in (a) indicate the presence of blood disease.}
\label{fig2} 
\end{figure}

\PARstart{H}{istopathology} images are considered to be the gold standard in the diagnosis of many diseases \cite{Gurcan2009histopathological}. In many situations, the cellular components are an important determinant. For example, in the biopsy sections of bone marrow, the abnormal cellular constitution indicates the presence of blood disease \cite{Bennett1976Proposals}. Bone marrow is the key component of both the hematopoietic system and the lymphatic system by producing large amounts of blood cells. The cell lines undergoing maturation in the marrow mostly include myeloid cells (granulocytes, monocytes, megakaryocytes, and their precursors), erythroid cells (normoblasts), and lymphoid cells (lymphocytes and their precursors). Figure~\ref{fig1} are examples of five main cellular components in bone marrow. These components are significant to both the systemic circulation and the immune system. Several kinds of cancer are characterized by the cellular constitution in bone marrow \cite{Bennett1976Proposals}. For instance, too many granulocytes precursors such as myeloblasts indicate the presence of chronic myeloid leukemia. Having large, abnormal lymphocytes heralds the presence of lymphoma. Figure~\ref{fig2} shows the difference between normal and abnormal bone marrow histopathology images from the perspective of cells. 

As described above, cell-level information is irreplaceable for histopathology image analysis. Cell-level visual attributes such as the morphological features of nuclei and the openness of chromatin are helpful for various tasks such as cell-level classification and nuclei segmentation. We define cell-level images as the output from nuclei segmentation. Each cell-level image contains only one cell. We opt to perform representation learning on these cell-level images, in which the visual attributes such as the nuclei morphology and chromatin openness are distinguished. The learned features are further utilized to assist tasks such as cell counting to highlight the quantification of certain types of cells.

To achieve this, the main obstacle is the labeling of cells. There are massive amounts of cells in each histopathology image, which makes manual labeling ambiguous and laborious. Therefore, an unsupervised cell-level visual representation learning method based on unlabeled data is believed to be more reasonable than fully supervised methods. Unsupervised cell-level visual representation learning is known to be difficult. First, geometrical and morphological appearances of cells from the same category can have a distinct diversity due to factors such as cell cycles. Furthermore, the staining conditions of histopathology images can be pretty diverse, resulting in inconsistent color characteristics of nuclei and cytoplasm.

Recently, deep learning has been proven to be powerful in histopathology image analysis such as classification \cite{Xu2014Deep,xu2015deep}, segmentation \cite{xu2016deep,Chen2016DCAN}, and detection \cite{chen2014deep,cirecsan2013mitosis}. Generative Adversarial Networks (GANs) \cite{goodfellow2014generative} are a class of generative models that use unlabeled data to perform representation learning. GAN is capable of transforming noise variables into visually appealing image samples by learning a model distribution that imitates the real data distribution. Several GAN architectures such as Deep Convolutional Generative Adversarial Nets (DCGAN) \cite{radford2015unsupervised} have proven their advantages in various natural images datasets. Recently, Wasserstein-GAN (WGAN) \cite{arjovsky2017wasserstein} and WGAN with gradient penalty (WGAN-GP) \cite{gulrajani2017improved} have greatly improved the stability of training GAN. More complex network structures such as residual networks \cite{he2016deep} can now be fused into GAN models.

Meanwhile, Information Maximizing Generative Adversarial Networks (InfoGAN) \cite{chen2016infoGAN} makes a modification that encourages GAN to learn interpretable and meaningful representations. InfoGAN maximizes the mutual information between the chosen random variables and the observations to make variables represent interpretable semantic features. The problem is that InfoGAN utilizes a DCGAN architecture, which requires meticulous attention towards hyperparameters. For our problem, it suffers a severe convergence problem.

Inspired by WGAN-GP and InfoGAN, we present an unsupervised representation learning method for cell-level images using a unified GAN architecture with a new formulation of loss, which inherits the superiority from both WGAN-GP and InfoGAN. We observe great improvements followed by the setting of WGAN-GP. Introducing mutual information into our formulation, we are capable of learning interpretable and disentangled cell-level visual representations, as well as allocate cells into different categories according to their most significant semantic features. Our method achieves promising results in the unsupervised classification of bone marrow cellular components.

Based on the cell-level visual representations, the quantification of each cellular component can be obtained by the trained model. Followed by this, cell proportions for each histopathology image can then be calculated to assist image-level classification. We further develop a pipeline combining cell-level unsupervised classification and nuclei segmentation to conduct image-level classification of histopathology images, which shows its advantages via experimentations on bone marrow datasets. 


The contributions of this work include the following: (1) We present an unsupervised framework to perform cell-level visual representation learning using generative adversarial networks. (2) A unified GAN architecture with a new formulation of loss is proposed to generate representations that are both high-quality and interpretable, which also endows our model the capability of cell-level unsupervised classification. (3) A pipeline is developed that exploits the varieties of cell-level elements to perform image-level classification of histopathology images.

\section{related works}

\subsection{Directly Related Works}
\subsubsection{Generative Adversarial Networks}

Goodfellow et al. \cite{goodfellow2014generative} propose GANs, a class of unsupervised generative models consisting of a generator neural network and an adversarial discriminator neural network. While the generator is encouraged to produce synthetic samples, the discriminator learns to discriminate between generated and real samples. This process is described as a minimax game. Radford et al. \cite{radford2015unsupervised} propose one of the most frequently used GAN architectures DCGAN.

Arjovsky et al. \cite{arjovsky2017wasserstein} propose WGAN, which modifies the objective function, securing the training process to be more stable. For regular GANs, the training process optimizes a lower bound of the Jensen-Shannon (JS) divergence between the generator distribution and the real data distribution. WGAN modifies this by optimizing an approximation of the Earth-Mover (EM) distance. The only challenge is how to enforce the Lipschitz constraint on the discriminator. While Arjovsky et al. \cite{arjovsky2017wasserstein} use weight-clipping, Gulrajani et al. \cite{gulrajani2017improved} propose WGAN-GP, which adds a gradient penalty on the discriminator. For our bone marrow datasets, even if we have tried multiple hyperparameters, DCGAN still suffers from a severe convergence difficulty. While DCGAN leads to the failure for our datasets, WGAN-GP greatly eases this problem.  

Chen et al. \cite{chen2016infoGAN} introduce mutual information into GAN architecture. Mutual information describes the dependencies between two separate variables. Maximizing mutual information between the chosen random variables and the generated samples, InfoGAN produces representations that are meaningful and interpretable. To exploit the varieties of cellular components, the superior ability of InfoGAN in learning disentangled and discrete representations is what a regular GAN lacks.

Therefore, we propose a unified GAN architecture with a new formulation of loss, which inherits the superiority of both WGAN-GP and InfoGAN. The outstanding stability of WGAN-GP eases the difficulty in tuning the complicated hyperparameters of InfoGAN. Introducing mutual information into our model, we are capable of learning interpretable cell-level visual representations, as well as allocate cells into different categories according to their most significant semantic features.



\subsubsection{Classification of Blood Disease}



Nazlibilek et al. \cite{Nazlibilek2014Automatic} propose a system to help automatically diagnose acute lymphocytic leukemia. This system consists of several stages: nuclei segmentation, feature extraction, cell-level classification, and cell counting. In their future work, they claim that the result of cell counting can be used for further diagnosis of acute lymphocytic leukemia.

In our work, we design a similar workflow which consists of nuclei segmentation, cell-level classification, and image-level classification. Our advantages lie in the novelty of an unsupervised setting and the convincing performance of image-level classification based on the calculated cell proportions.

\subsection {Cell-level Representation} 

The representation of individual cells can be used for a variety of tasks such as cell classification. Traditional cell-level visual representation for classification tasks can be categorized into four categories \cite{Y2014Methods}: morphological \cite{Muthu2012Hybrid}, texture \cite{Xu2015Dual,Lorenzo2013Cervical}, intensity \cite{Dundar2011Computerized}, and cytology features \cite{Nguyen2011Prostate}. These traditional methods have been employed in the representation of white blood cells \cite{Tai2011Blood,Putzu2014Leucocyte,Su2014A}. However, the features used above need to be manually designed by experienced experts according to the characteristics of different types of cells. While images suffer from a distinct variance, discovering, characterizing and selecting good handcraft features can be extremely difficult.

To remedy the limitations of manual features in cell classification, Convolutional Neural Network (CNN) learns higher-level latent features, whose convolution layer can act as a feature extractor \cite{xu2017large}. Xie et al. \cite{xie2016unsupervised} propose Deep Embedding Clustering (DEC) that simultaneously learns feature representations and cluster assignments using deep neural networks.

Variational Autoencoder (VAE) \cite{kingma2013auto} serves as a convincing unsupervised strategy in cell-level visual representation learning \cite{Xu2016Stacked,Cruzroa2013A, zhang2016fusing}. However, how to use VAE to learn categorical and discrete latent variables is still under investigation. Dilokthanakul et al. \cite{dilokthanakul2016deep} and Jiang et al. \cite{jiang2017variational} design models combining VAE with Gaussian Mixture Model (GMM). But they demonstrate their experiment on one-dimensional datasets such as MNIST. To perform clustering and embedding on a higher-dimensional dataset, their methods still need a feature extractor. 

GANs such as Categorical GAN \cite{springenberg2015unsupervised} can merge categorical variables into the model with little effort, which makes learned representations disentangled and interpretable. This ability is critical in medical image analysis where accountability is especially needed.

\subsection {Cell-level Histopathology Image Analysis} 
\subsubsection {Classification} 

Cell classification has been performed in diverse histopathology related works such as breast cancer \cite{Malon2013Classification}, acute lymphocytes leukemia \cite{Mohapatra2014An,Zhao2016Automatic}, and colon cancer \cite{Sirinukunwattana2016Locality}.

Based on the result of cell classification, some approaches have been proposed to determine the presence or location of cancer \cite{Nguyen2011Prostate}, \cite{hou2016automatic}. In prostate cancer, Nguyen et al. \cite{Nguyen2011Prostate} innovatively employ cell classification for automatic cancer detection and grading. They distinguish the cancer nuclei and normal nuclei, which are combined with textural features to classify the image as normal or cancerous and then detect and grade the cancer regions. In the diagnosis of Glioma, Hou et al. \cite{hou2016automatic} apply CNN to the classification of morphological attributes of nuclei. They also claim that the nuclei classification result provides clinical information for diagnosing and classifying glioma into subtypes and grades. {Zhang et al. \cite{zhang2015weighted, zhang2015towards, zhang2015high} and Shi et al. \cite{shi2017cell} use either supervised or semi-supervised hashing models for cell-level analysis.}

All of these works require a large amount of accurately annotated data. Obtaining such annotated data is time-consuming and labor-intensive while GAN can optimally leverage the wealth of unlabeled data. 

\subsubsection{Segmentation}
Nuclei segmentation is of great importance for cell-level classification. Nuclei segmentation methods can be roughly categorized as follows: intensity thresholding \cite{Callau2015Evaluation}, \cite{Wienert2012Detection}, morphology operation \cite{Dorini2013Semiautomatic}, \cite{Schmitt2009Morphological}, deformable models \cite{Dzyubachyk2010Advanced}, watershed transform \cite{Long2009A}, clustering \cite{Hai2014Automatic}, \cite{Bueno2012A}, and graph-based methods \cite{Chang2013Invariant}, \cite{Arslan2013Attributed}. The methods above have been broadly applied to the segmentation of white blood cells.

\subsection {Generative Adversarial Networks in Medical Images}
Recently, several works involving GAN have gathered great attention in medical image analysis. 

In medical image synthesizing, Nie et al. \cite{nie2017medical} estimate the CT image from its corresponding MR image with context-aware GAN. In medical image reconstruction, Li et al. \cite{Li2017Reconstruction} use GAN to reconstruct medical images with the thinner sliced thickness from regular thick-slice images. Mahapatra et al. \cite{Mahapatra2017Image} propose a super resolution method that takes a low-resolution input fundus image to generate a high-resolution super-resolved image. Wolterink et al. \cite{Wolterink2017Generative} employ GAN to reduce the noise in low-dose CT images. All these recent works demonstrate the great potential of GAN in solving complicated medical problems.

\section{Methods}

In this section, we first introduce an unsupervised method for cell-level visual representation learning using GAN. Then we present the details of how image-level classification is performed on histopathology images based on cell-level representation. 

\begin{figure}[H]
  \subfigure[Training process. Random variables are composed of Gaussian variables $z$ and the discrete variable $c$. Besides playing the minimax game between the generator ($G$) and the discriminator ($D$) through the EM distance, we also minimize the negative Log-likelihood between $c$ and the output of the auxiliary network ($Q(c|G(c,z)$) to maximize mutual information.]{ 
    \label{fig:mini:subfig:train}
    \begin{minipage}[b]{0.48\textwidth} 
      \centering 
      \includegraphics[width=1\textwidth]{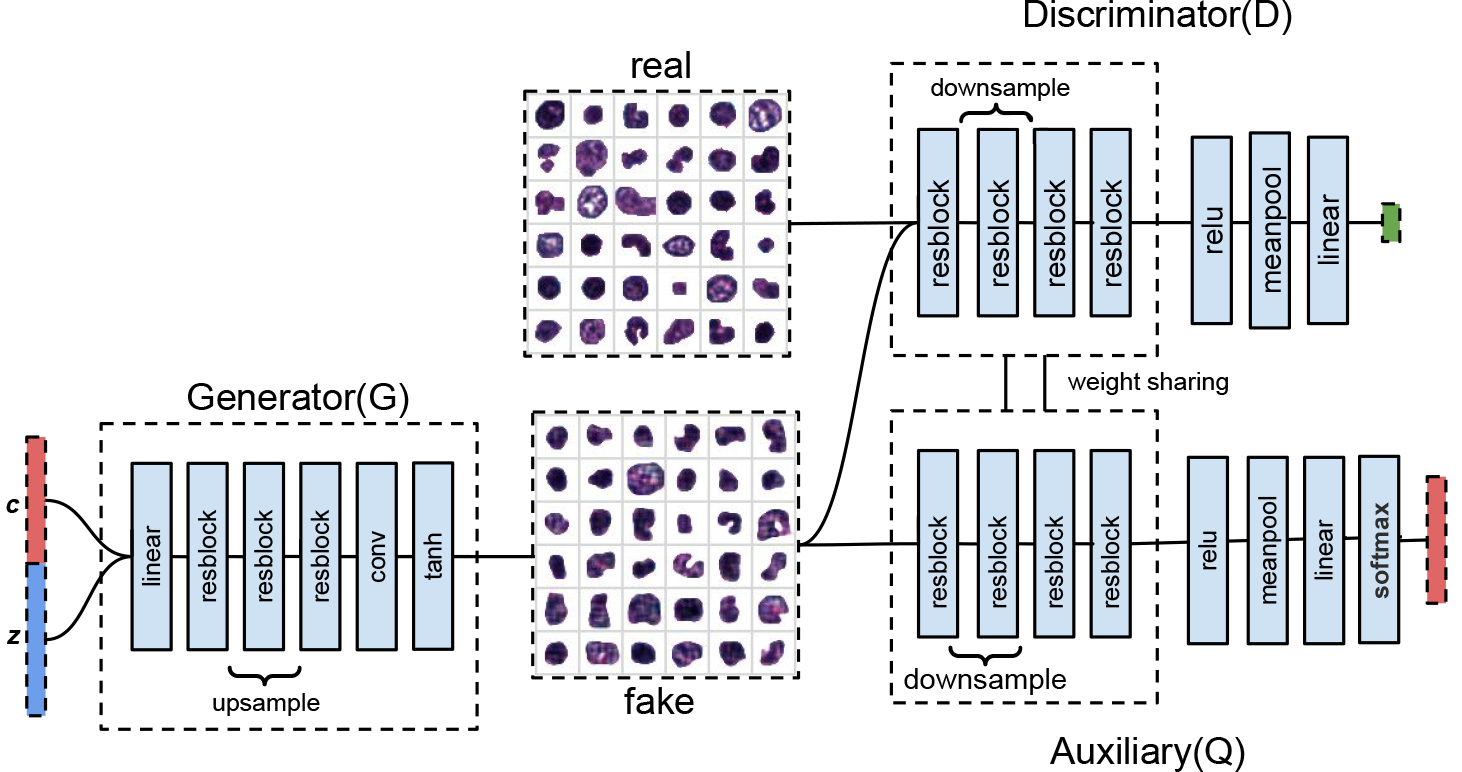} 
    \end{minipage}
    }
  \subfigure[Test process. Real samples are classified into five categories by the auxiliary network $Q$. At the same time, fake samples are generated by giving noises with the chosen $c$ for each class. In the example of generated samples (fake), one row contains five samples from the same category in $c$, and a column shows the generated images for 5 possible categories in $c$ with $z$ fixed.]{
    \label{fig:mini:subfig:b}
    \begin{minipage}[b]{0.48\textwidth} 
      \centering
      \includegraphics[width=0.95\textwidth]{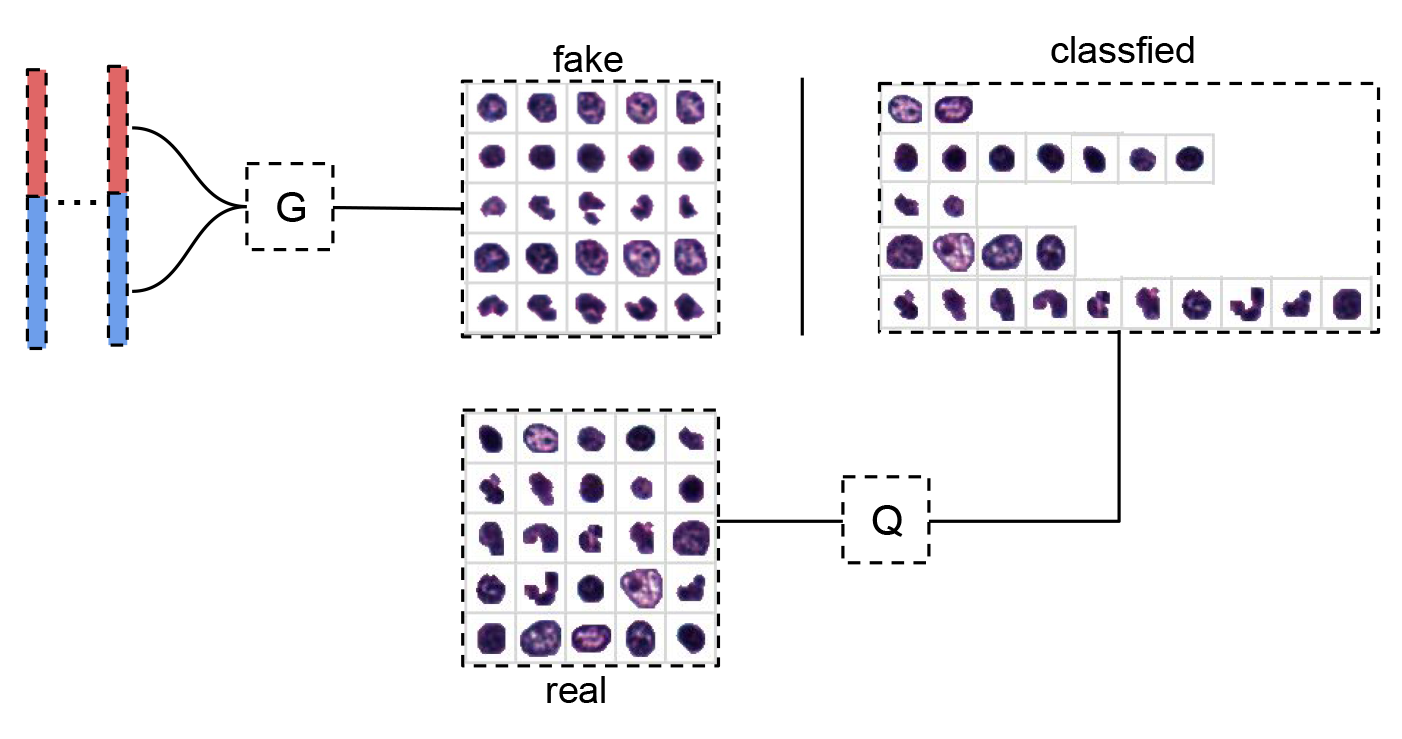} 
    \end{minipage}}
  \subfigure[Illustration of residual blocks (resblocks) in the architecture. There are three different types of residual blocks considering whether they include nearest-neighbor upsampling or mean pooling for downsampling. Batch normalization layers are used in our generator to help stabilize training.]{ 
    \label{fig:mini:subfig:c}
    \begin{minipage}[b]{0.48\textwidth} 
      \centering 
      \includegraphics[width=0.95\textwidth]{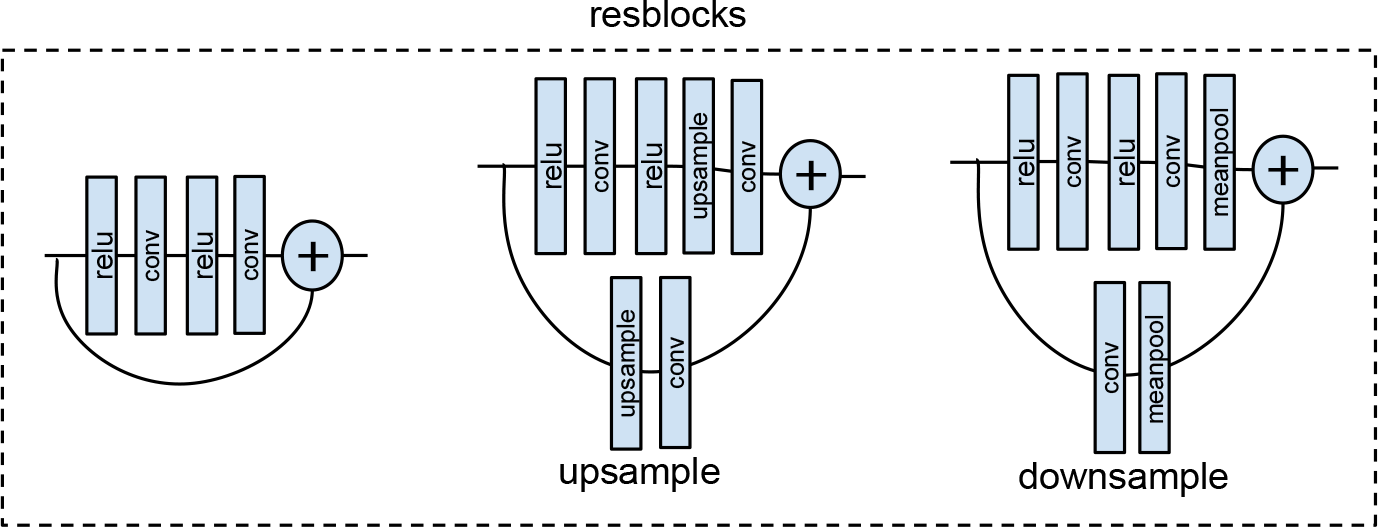} 
    \end{minipage}} 
  \caption{Network architecture of our cell-level visual representation learning: (a) Training process. (b) Test process. (c) The architecture of residual blocks (written as resblock in (a) and (b)).} 
  \label{networks architecture} 
\end{figure}

\subsection{Cell-level Visual Representation Learning}

Given cell-level images that come from nuclei segmentation as the real data, we define a generator network $G$, a discriminator network $D$, and an auxiliary network $Q$. The architecture of these networks are shown in Figure~\ref{networks architecture}. In the training process, we learn a generator distribution that matches the real data distribution by playing a minimax game between $G$ and $D$ by optimizing an approximation of the Earth-Mover (EM) distance. Meanwhile, we maximize mutual information between the chosen random variables and the generated samples using an auxiliary network $Q$. In the test process, the generator generates the representations for each category of cells according to different values of the chosen random variables. Cell images can be allocated to the corresponding categories by the auxiliary network $Q$.

\begin{figure*}[t]
\includegraphics[width=1\textwidth]{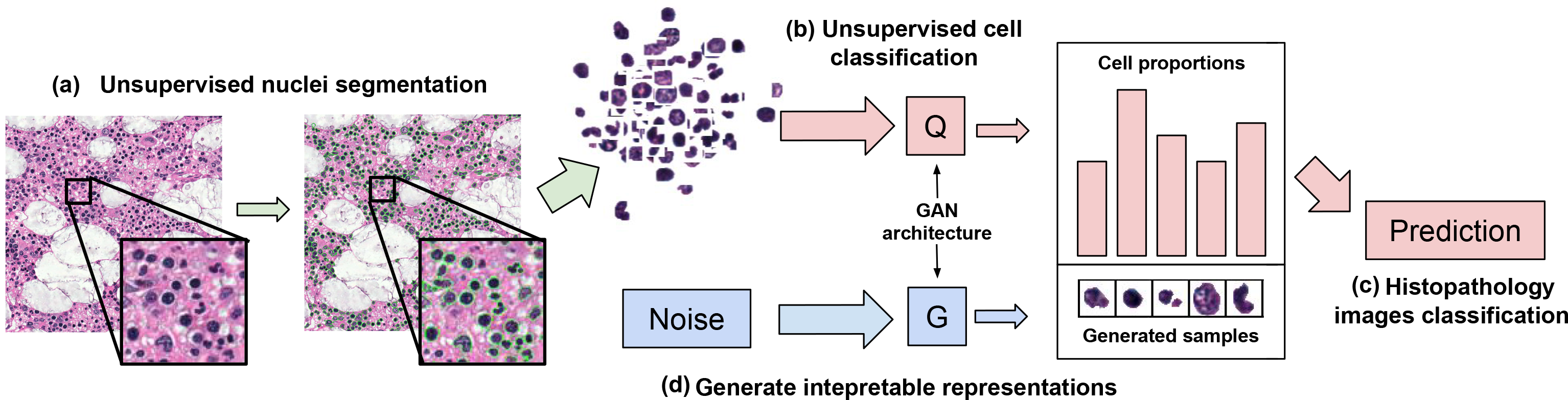} \caption{Overview of our pipeline as follows: (a) Nuclei segmentation is performed on histopathology images. (b) Using the trained GAN architecture, Cell-level clustering is performed using the learned auxiliary network $Q$. Cell proportions are then calculated for each histopathology image. (c) Image-level prediction is given based on cell proportions. (d) For visualization, the generator $G$ can generate the interpretable representation for each category of cells by changing the noises.} 
\label{pipeline} 
\end{figure*}

\subsubsection {Training Process}
Given cell-level images sampled from the real data distribution $x\sim \mathbb{P}_r$, the first goal is to learn a generator distribution $\mathbb{P}_g$ that matches the real data distribution $\mathbb{P}_r$.

We first define a random noise variable $z$. The input noise $z$ is transformed by the generator into a sample $\tilde x = G(z), z \sim p(z)$. $\tilde x$ can be viewed as following the generator distribution $\mathbb{P}_g$. Inspired by WGAN \cite{arjovsky2017wasserstein}, we optimize networks through the WGAN objective $W(\mathbb{P}_r, \mathbb{P}_g)$:
\begin{equation}
W(\mathbb{P}_r, \mathbb{P}_g) = \sup_{\lVert f \lVert_{L \leq 1}} \ \mathbb{E}_{x \sim \mathbb{P}_r}[f(x)] - \mathbb{E}_{\tilde x \sim \mathbb{P}_g}[f(\tilde x)].
\end{equation}

$W(\mathbb{P}_r, \mathbb{P}_g)$ is an efficient approximation of the EM distance, which is constructed using the Kantorovich-Rubinstein duality \cite{arjovsky2017wasserstein}. The EM distance measures how close the generator distribution and the data distribution are. To distinguish two distributions $\mathbb{P}_g$ and $\mathbb{P}_r$, the adversarial discriminator network $D$ is trained to learn the function $f$ that maximizes $W(\mathbb{P}_r, \mathbb{P}_g)$. To make $\mathbb{P}_g$ approach $\mathbb{P}_r$, the generator instead is trained to minimize $W(\mathbb{P}_r, \mathbb{P}_g)$. The value function $V(D,G)$ is written as follows:
\begin{equation}
V(D,G) = \mathbb{E}_{x \sim \mathbb{P}_r} [D(x)] - \mathbb{E}_{z \sim p(z)} [D(G(z))].
\end{equation} This minimax game between the generator and the discriminator is written as: 
\begin{equation}
\min_{G}\max_{D \in \mathcal{D}} V(D,G).
\end{equation}

Followed by the work of WGAN-GP \cite{gulrajani2017improved}, a gradient penalty is added on the discriminator to enforce the Lipschitz constraint to make sure that the discriminator lies within the space of 1-Lipschitz functions $D\in\mathcal{D}$. The loss of the discriminator with a hyperparameter $\lambda_1$ is written as:
\begin{equation}
\resizebox{1\hsize}{!}{$L_D = \mathbb{E}_{z \sim p(z)} [D(G(z))] - \mathbb{E}_{x \sim \mathbb{P}_r} [D(x)]+ \lambda_1 \mathbb{E}_{\hat x \sim \mathbb{P}_{\hat x}} [ || \nabla_{\hat x} D(\hat x) ||_p - 1 ]^2,$}
\end{equation} where $\mathbb{P}_{\hat x}$ is defined sampling uniformly along straight lines between pairs of points sampled from the data distribution $\mathbb{P}_r$ and the generator distribution $\mathbb{P}_g$. 

In this way, our model is capable of generating visually appealing cell-level images. But still, it fails to exploit information of categories of cells since the noise variable $z$ doesn't correspond to any interpretable feature. Motivated by this, our second goal is to make the chosen variables represent meaningful and interpretable semantic features of cells. Inspired by InfoGAN \cite{chen2016infoGAN}, we introduce mutual information into our model:
\begin{equation}
I(X;Y) = \mathrm {H} (X)-\mathrm {H} (X|Y) = \mathrm {H} (Y) -\mathrm {H} (Y|X).
\label{mutual}
\end{equation}

$I(X;Y)$ describes the dependencies between two separate variables $X$ and $Y$. It measures the different aspects of the association between two random variables. If the chosen random variables correspond to certain semantic features, it's reasonable to assume that mutual information between generated samples and random variables should be high.

We define a latent variable $c$ sampled from a fixed noise distribution $p(c)$. The concatenation of the random noise variable $z$ and the latent variable $c$ is then transformed by the generator G into a sample $G(z,c)$. Since we encourage the latent variable to correspond with meaningful semantic features, there should be high mutual information between $c$ and $G(z, c)$. Therefore, the next step is to maximize mutual information $I(c;G(z,c))$, which can be written as:
\begin{equation}
I(c;G(z,c)) =  H(c) - H(c \vert G(z,c)).
\end{equation} Followed by this, a lower bound $L_I$ is given by:
\begin{equation}
L_I(G, Q) = \mathbb{E}_{z \sim p(z), c \sim p(c)}[\log Q(c \vert G(z,c))] + H(c), 
\end{equation} where $H(c)$ is the entropy of the variable sampled from a fixed noise distribution. Maximizing this lower bound, we maximize mutual information $I(c;G(z,c))$. The proof can be found in InfoGAN \cite{chen2016infoGAN}.

\begin{figure}[h!]
\centering
\includegraphics[width=0.46\textwidth]{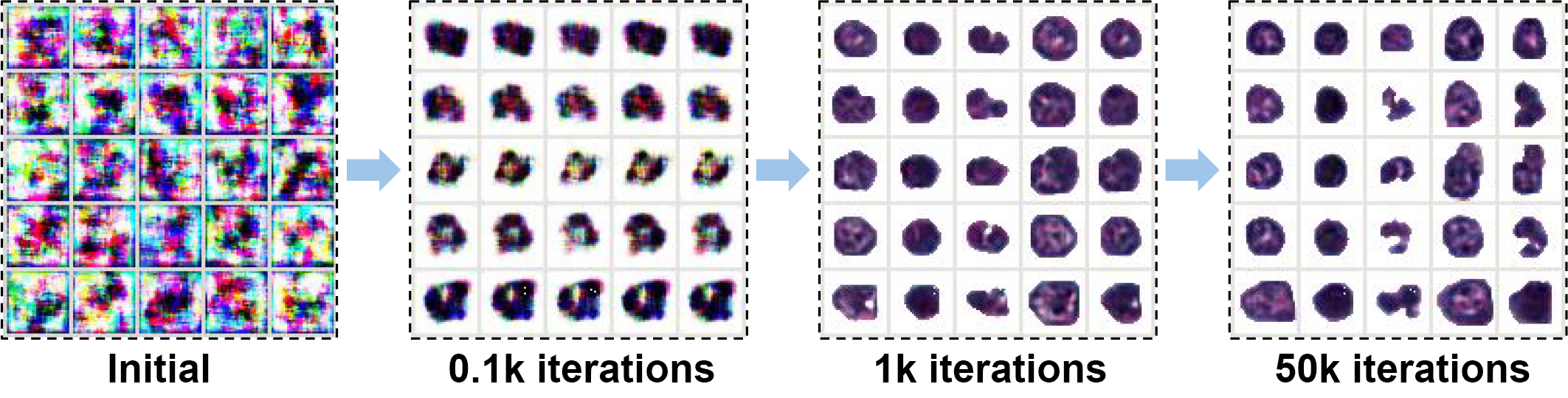}
\caption{Example of how a set of noise vectors are transformed into interpretable image samples over generator iterations. We use a 5-dimensional categorical variable $c$ and 32 Gaussian noise variables $z$ as input. Different rows correspond to different values of $z$. Different columns correspond to different values of $c$. The value of $c$ largely corresponds to cell types.}
\label{trainingpicture}
\end{figure}

Since we introduce the latent variable $c$ into the model, the value function $V(D,G)$ is replaced by:
\begin{equation}
V (D,G) \leftarrow \mathbb{E}_{x \sim \mathbb{P}_r} [D(x)] - \mathbb{E}_{z \sim p(z), c \sim p(c)}[D(G(z,c))].
\label{fig:wgan}
\end{equation}As we combine the adversarial process with the process of maximizing mutual information, this information-regularized minimax game with a hyperparameter $\lambda_2$ can be written as follows: 
\begin{equation}
\min_{G,Q}\max_{D \in \mathcal{D}}V (D,G)-\lambda_2 L_I(G, Q).
\end{equation} The loss of $D$ can be replaced by:
\begin{equation}
\resizebox{1\hsize}{!}{$L_D \leftarrow \mathbb{E}_{z \sim p(z), c \sim p(c)}[D(G(z,c))] - \mathbb{E}_{x\sim \mathbb{P}_r}[D(x)] + \lambda_1 \mathbb{E}_{\hat x \sim \mathbb{P}_{\hat x}} [ || \nabla_{\hat x} D(\hat x) ||_p - 1 ]^2,$}
\end{equation} Since $H(c)$ can be viewed as a constant, the loss of the auxiliary network $Q$ can be written as the negative log-likelihood between $Q(c|G(c,z))$ and the discrete variable $c$. The losses of $G$ and $Q$ can be interpreted as below:
\begin{equation}
L_G = -\mathbb{E}_{z \sim p(z), c \sim p(c)}[D(G(z,c))],
\end{equation}
\begin{equation}
L_Q = - \lambda_2 \mathbb{E}_{z \sim p(z), c \sim p(c)}[\log Q(c|G(z,c))].
\label{fig:InfoGAN}
\end{equation} Figure~\ref{trainingpicture} shows how noises are transformed into interpretable samples during the training process.

\subsubsection {Test Process}
In the training process, a generator distribution is learned to imitate the real data distribution. An auxiliary distribution is learned to maximize the lower bound. Especially if $c$ is sampled from a categorical distribution, a softmax function is applied as the final layer of $Q$. Under this circumstance, $Q$ can act as a classifier in the test process, since the posterior $Q(c|x)$ is discrete. Assuming that each category in $c$ corresponds to a type of cells, the auxiliary network $Q$ can divide cell-level images into different categories while the generator $G$ can generate the interpretable representation for each category of cells.

\subsection{Image-level Classification}
Based on the cell-level visual representation learning, we propose a pipeline combining nuclei segmentation and cell-level visual representation to highlight the varieties of cellular elements. Image-level classification is performed using the calculated cell proportions. The illustration of this pipeline is shown in Figure~\ref{pipeline}.

\subsubsection{Nuclei Segmentation}

\begin{figure}[t]
\includegraphics[width=0.48\textwidth]{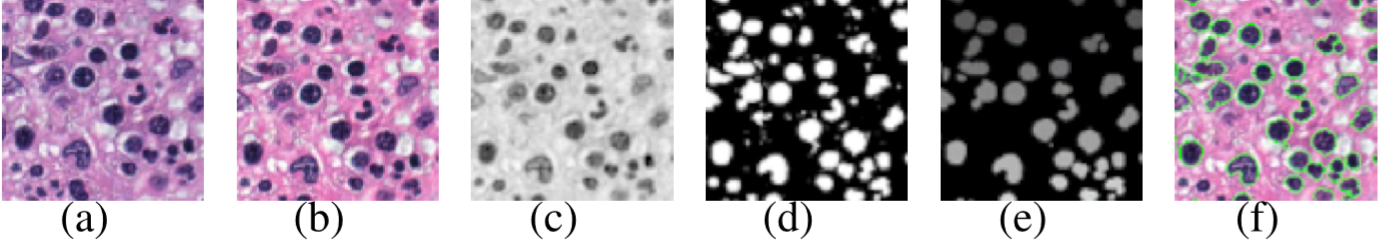}  
\caption{Overview of segementation process: $\displaystyle (a)$ the cropped image, $\displaystyle (b)$ the normalized image, $\displaystyle (c)$ the separated hematoxylin stain image using color deconvolution, $\displaystyle (d)$ the binary image generated by intensity thresholding, $\displaystyle (e)$ the labeled image after postprocessing where different grayscale values stand for different segmented instances, and $\displaystyle (f)$ the final segmentation image.} 
\label{segmentation} 
\end{figure}

An unsupervised nuclei segmentation approach is ultilized consisting of four stages: normalization, unsupervised color deconvolution, intensity thresholding and postprocessing to segment nuclei from the background. Figure~\ref{segmentation} is an overview of our segmentation pipeline.

\textbf{Color Normalization:} We employ Reinhard color normalization \cite{Reinhard2001Color} to convert the color characteristics of all images into the desired standard by computing the mean and standard deviations of a target image in LAB space.

\textbf{Color Deconvolution:} Using the PCA-based `Macenko' method \cite{Macenko2009A}, unsupervised color deconvolution is performed to separate the normalized image into two stains. We project pixels onto a best-fit plane, wherein it selects the stain vectors as percentiles in the `angle distribution' of the corresponding plane. With the correct stain matrix for color deconvolution, the normalized image can be separated into hematoxylin stain and eosin stain.

\textbf{Intensity Thresholding:} To sufficiently segment cells, we apply intensity thresholding in the hematoxylin stain image where the intensity distribution of cells is consistently distinct from the background. By converting the hematoxylin stain image into a binary image with a constant global threshold, the cells are roughly segmented.

\textbf{Postprocessing:} In image postprocessing, objects with fewer pixels than the minimum area threshold will be removed from the binary image. Then we employ the method in \cite{Wienert2012Detection} to remove thin protrusions from cells. Furthermore, we use opening operation to separate a few touched cells. 

\subsubsection{Classification}
We utilize the model distribution trained in our unsupervised representation learning as the cell-level classifier. Assuming that we use a $k$-dimensional categorical variable as the chosen variable in the training process, the real data (cell-level images) distribution is allocated into $k$ dimensions. In the test process, cell-level images are unsupervised classified into $k$ corresponding categories. 

For each histopathology image, we count the numbers of cell-level instances in each category as the representation of its cellular constitution, denoted as $\{X_1, X_2, X_3, \ldots, X_k\}$. For cellular element $i$, the ratio of the number of this cellular element to the total number of the cellular constitution in this image is calculated by $P_i=\frac{X_i}{\sum_{i=1}^{k}{X_i}}$. We define $P_i$ as the cell proportion of cellular element $i$. 

Given cell proportions $\{P_1, P_2, P_3, \ldots, P_k\}$ as the feature vector of histopathology images, we utilize either k-means or SVM to give image-level predictions. 

\section{Experiments and Results}

\subsection{Dataset} \label{subsection:dataset}
All our experiments are conducted on bone marrow histopathology images stained with hematoxylin and eosin. As described before, the cellular constitution in bone marrow is a determinant in diagnoses of blood disease.

\textbf{Dataset A:} Publicly available dataset \cite{kainz2015you} which consists of eleven images of healthy bone marrow with a resolution of $1200 \times 1200$ pixels. Each image contains around 200 cells. The whole dataset includes 1995 cell-level images in total. We label all cell-level images into four categories: 34 neutrophils, 751 myeloblasts, 495 monocytes, and 715 lymphocytes. Images are carefully labeled by two pathologists. When the two pathologists disagree on a particular image, a senior pathologist makes a decision over the discord.

\textbf{Dataset B:} Dataset provided by the First Affiliated Hospital of Zhejiang University which contains whole slides of bone marrow from 24 patients with blood diseases. Each patient matchs with one whole slide. We randomly crop 29 images with a resolution of $1500 \times 800$ pixels from all whole slides. Dataset B contains around 12000 cells in total. For this dataset, we label 600 cell-level images into three categories for evaluation: 200 myeloblasts, 200 monocytes, and 200 lymphocytes. The labeling process is conducted in the same manner as Dataset A.

\textbf{Dataset C:} Combination of Datasets A and B, which results in 29 abnormal and 11 normal histopathology images. 

\textbf{Dataset D:} Dataset includes whole slides from 28 patients with bone marrow hematopoietic tissue hyperplasia (negative) and 56 patients with leukemia (positive). Each patient matchs with one whole slide. We randomly crop images with a resolution of $1500 \times 800$ pixels from all whole slides. This results in 72 negative and 132 positive images. After segmentation, Dataset D contains around 80000 cells in total.
\subsection{Implementation}

\textbf{Network Parameters:} Our generator $G$, discriminator $D$ and auxiliary network $Q$ all have the structures of residual networks. In the training process, all three networks are updated by Adam optimizer ($\alpha=0.0001$, $\beta_1=0.5$, $\beta_2=0.9$, $lr=2 \times 10^{-4}$) \cite{kingma2014adam} with a batch size of 64. All our experiments use hyperparameters $\lambda_1 = 10$ and $\lambda _2 = 1$. For each training iteration, we update $D$, $G$ and $Q$ in turn. One training iteration consists of five discriminator iterations, one generator iteration, and one auxiliary network iteration. For each training process, we augment the training set by rotating images with angles $90^\circ$, $180^\circ$, $270^\circ$. We train ten epochs for our model in each experiment.

\textbf{Noise Sources:} The noise fed into the network is the combination of a 5-dimensional categorical variable and 32 Gaussian noise variables for the training of Dataset A or Dataset B. We use the combination of a 5-dimensional categorical variable and 64 Gaussian noise variables for Dataset C.

\textbf{Segmentation Parameters:} The mean value of the standard image in three channels is $[8.98\pm 0.64, 0.08\pm 0.11, 0.02\pm 0.03]$ for color normalization. Vectors for color deconvolution are picked from 1\% to 99\% angle distribution while the magnitude below 16 is excluded from the computation. We use the threshold value of 120 for intensity thresholding. In the post-process, objects with pixels smaller than 200 will be removed. An opening operation with $7 \times 7$ kernel size is performed to separate touched cells. When the edge of the bounding box of a cell-level image is larger than 32 pixels, we rescale the image to make the larger edge match to 32. Each cell is centered in a $32 \times 32$ pixel image where blank is filled with $[255, 255, 255]$.

\textbf{Bounding Box:} To prevent the color and texture contrast from troubling the feature extraction process, we use instances without segmentation for baseline methods. If we depose the nuclei in the center with the loose bounding box in the same manner as our previous experiments, cells will suffer from severe overlapping. Thus, we crop the minimum bounding box region along each segmented instance, and then resize it into $32 \times 32$ pixels as our dataset.

\textbf{Software:} We implement our experiments on framework Pytorch for deep learning models and framework HistomicsTK for nuclei segmentation. Our model is compared with multiple sources of baselines. Three main types of baselines are claimed to be relevant as follows: (1) feature extractors including manual features, HOG and DNN extractor; (2) supervised classifiers including SVM and DNN; (3) clustering algorithms including DEC and K-means. The rich mix of different sources of baselines, including deep learning algorithms, provides a stronger demonstration to our experiments. We utilize k-means++ \cite{arthur2007k} to choose the initial values when using k-means to perform clustering. The feature code\footnote{Implementation details can be found at \url{https://github.com/bohu615/nu_gan}} is Python implementation in all these algorithms.

\textbf{Hardware:} For hardware, we use one pair of Tesla K80 GPU for parallel training and testing of neural network models. Other baseline experiments are conducted on Intel(R) Xeon(R) CPU E5-2690 v3 @ 2.60GHz. For our model, with a batch size of 64, using one pair of K80 GPU for parallel computation, each generator iteration costs 3.2 seconds in the training process when each batch costs 0.18 seconds in the test process.
\subsection{Cell-level Classification Using Various Features}

To demonstrate the quality of our representation learning, we apply the trained model as a feature extractor. The experiment is conducted on Dataset A. In this experiment, 1596 cell-level images are used for training; 399 cell-level images are used for testing. 

\textbf{Comparison:} (1) MF: 188-dimensional manual feature combined of SIFT \cite{lowe2004distinctive}, LBP \cite{ojala2002multiresolution}, and $L \times a \times b$ color histogram. (2) DNN: DNN+k-means: DNN features extracted by ResNet-50 trained on Imagenet-1K, on top of which k-means is performed. (3) Our Method: We downsample the features after each residual block of the discriminator into a $4 \times 4$ spatial grid using max pooling. These features are flattened and concatenated to form an 8192-dimensional vector. On top of the feature vectors, an L2-SVM is trained to perform classification. 

Different processing strategies are used as follows: (1) w/ Seg: using the output generated by nuclei segmentation; (2) w/o Seg: using the minimum bounding box along each cell-level instance. 

\textbf{Evaluation:} For each class, we denote the number of true positives $TP$, the number of false positives $FP$ and the number of false negatives $FN$. The precision, recall and F-score ($F_1$) for each class are defined as follows: \begin{equation}
\begin{aligned}
& precision=\frac{TP}{TP+FP}, \\
& recall=\frac{TP}{TP+FN}, \\
& F_1 = \frac{2 \cdot precision \cdot recall}{precision + recall}. \\
\end{aligned}
\label{eq:fscore}
\end{equation}The average precision, recall and F-socre are calculated weighted by support (the number of true instances of each class).

\begin{figure}[H]
\centering
  \includegraphics[width=0.38\textwidth]{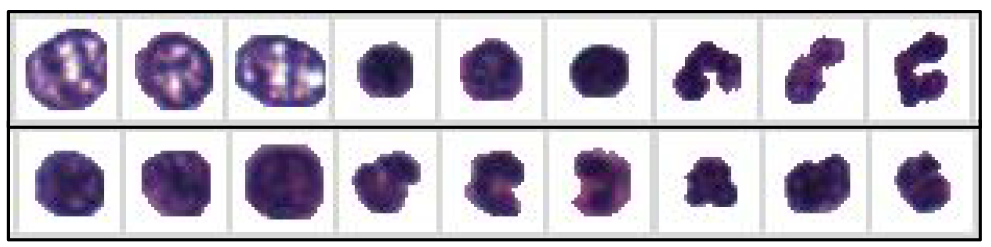} 
  \caption{Visualization of cell-level classification performed on Dataset A: $(up)$ correctly classified samples and $(down)$ misclassified samples. misclassified samples can be illegible for pathologists either.} 
  \label{cell_right_wrong} 
\end{figure}

\textbf{Results:} We randomly choose correctly classified and misclassified samples displayed in Figure~\ref{cell_right_wrong}. The comparison of results is shown as Table~\ref{cell-level classification}, which proves the advantages of our representation learning method. The manual feature extractor can generate a better result based on the bounding box regions, but its performance is still lower than ours. The color of the background can provide useful information for the color histogram channel in manual features but is viewed as noise for the DNN based extractor. Though the dimensions of the feature vectors of our method are higher, the clustering ability of our model ensures further unsupervised applications. Furthermore, we apply mean pooling on top of feature maps to prove that using less dimensional features can also generate a comparable result. In this manner, we achieve 0.850 F-score using 2048 dimensional features and 0.840 F-score using 512 dimensional features.

\begin{table}[H]
\centering
\caption{{\textnormal{Performance of cell-level classification using various features.}}}
\label{cell-level classification}
\begin{tabular}{|c|c|c|c|c|c|c|}
\hline
\multirow{2}{*}{Methods} & \multicolumn{2}{c|}{Precision} & \multicolumn{2}{c|}{Recall} & \multicolumn{2}{c|}{F-score} \\ \cline{2-7} 
                         & \tiny{w/ Seg}            & \tiny{{w/o Seg}}    & \tiny{{w/ Seg}}           & \tiny{{w/o Seg}}  & \tiny{{w/ Seg}}           & \tiny{{w/o Seg}}   \\ \hline
MF                       & 0.821             & {0.837}      & 0.803            & {0.847}    & 0.811                 & {0.842}     \\
DNN                      & 0.838             & {0.760}      & 0.817            & {0.769}    & 0.827                 & {0.764}     \\
Our Method               & \textbf{0.865}    & {/}          & \textbf{0.848}   & {/}        & \textbf{0.857}        & {/}         \\ \hline
\end{tabular}
\end{table}

\subsection{Cell-level Clustering}

As the priority of image-level classification of histopathology images, cell-level clustering is performed using the trained auxiliary network $Q$. We conduct experiments on the three datasets described in Section~\ref{subsection:dataset}.

\textbf{Comparison:} (1) MF+k-means: Manual features with k-means. (2) DNN+k-means: DNN features extracted by ResNet-50 trained on Imagenet-1K, on top of which k-means is performed. (3) HOG+DEC: Deep Embedded Clustering (DEC) \cite{xie2016unsupervised} on 2048-dimensional HOG features. (4) Our Method: Cell images are unsupervised allocated to five clusters by the auxiliary network $Q$. We also test models such as Categorical GAN (CatGAN) \cite{springenberg2015unsupervised}, InfoGAN (under DCGAN architecture), and Gaussian Mixture VAE (GMVAE) \cite{dilokthanakul2016deep} on our datasets under different hyperparameters, but find them fail to converge.

{The following processing strategies are also used: (1) w/ Seg: using the output generated by nuclei segmentation; (2) w/o Seg: using the minimum bounding box along each cell-level instance.}

\textbf{Evaluation:} We evaluate the performance of clustering using the average F-score, purity, and entropy. For the set of clusters $\{ \omega_1, \omega_2, \ldots, \omega_K \}$ and the set of classes $\{ c_1,c_2,\ldots,c_J \}$, we assume that each cluster $\omega_k$ is assigned to only one class ${\mathop{\mathrm{argmax}}\nolimits}_j (\vert\omega_k \cap c_j\vert)$. The F-score for class $c_j$ is then given by Equation \ref{eq:fscore}. The average F-score is given calculated by the number of true instances in each class.

Purity and Entropy are also used as evaluation metrics, which are written as follows:
\begin{equation}
\begin{aligned}
& purity = \frac{1}{N} \sum_k \max_j \vert\omega_k \cap c_j\vert, \\
& entropy = -\frac{1}{N}\sum_k\vert \omega_k\vert \log \frac{\vert \omega_k\vert}{N}. \\
\end{aligned}
\label{eq:purity}
\end{equation}Larger purity and smaller entropy indicate better clustering results.

For nuclei segmentation, we use Intersection over Union (IoU) and the F-score as evaluation metrics. A segmented instance (I) is matched with the ground truth (G) only if they intersect at least 50\% (i.e., $\vert I \cap G \vert>0.5G$). For each matched instance and its ground truth, the overlapping pixels are counted as true positive ($TP$). The pixels of instance remain unmatched are counted as false positive ($FP$) while the pixels of ground truth remaining unmatched are counted as false negative ($FN$). The F-score is then calculated using Equation \ref{eq:fscore}.

For k-means based methods, the average F-score is approximately the same ($\pm 0.02$) using either four, five, or six clusters. 

\begin{figure*}[t]
\centering
\includegraphics[width=\textwidth]{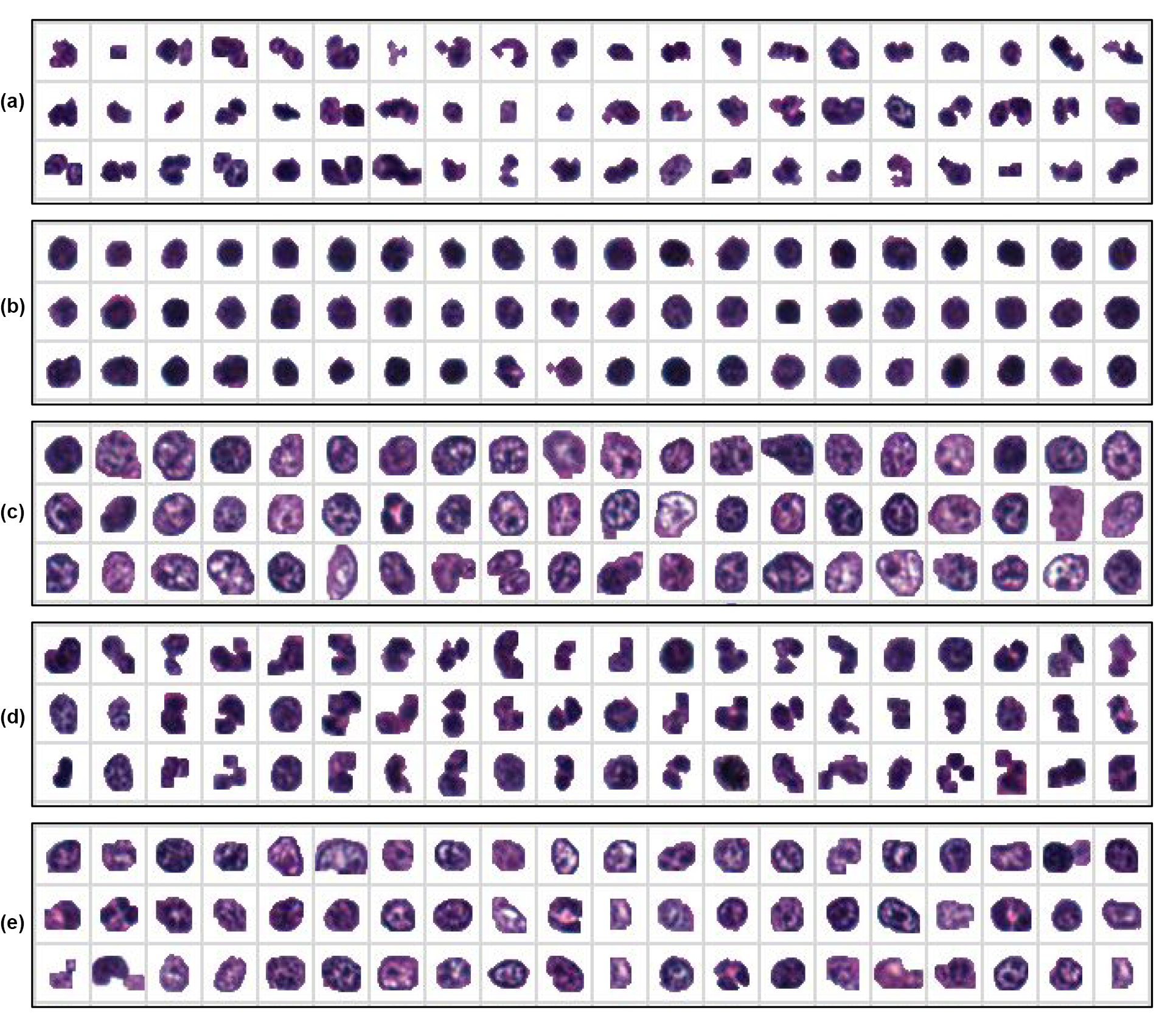}
\caption{Visualization of clustering. We randomly select 60 samples from each one of five clusters, displayed as (a) to (e). Instances in the same cluster have a distinct consistency. In (b), cells in marrow with dark, dense, and close phased nuclei tend to be lymphocytes or erythroid precursors. In (c) and (e), cells with dispersed chromatin are most likely granulocytes precursors such as myeloblasts.}
\label{cluster_display}
\end{figure*}

\textbf{Annotations:} To evaluate the capability of nuclei segmentation, We randomly choose 20 patches from Dataset C with a resolution of $200 \times 200$ pixels. The ground truth is carefully labeled by two pathologists. When the two pathologists disagree on a particular image, a senior pathologist makes a decision over the discord. 

\textbf{Results:} For nuclei segmentation, our method achieves 0.56 mean IoU and 0.70 F-score. 

For cell-level clustering, the comparison shown as Table~\ref{table:cluster} shows the superiority of our method. To explicitly reveal the semantic features our model has captured, we randomly choose 60 samples from each of the five clusters displayed in Figure~\ref{cluster_display}, which shows a distinct consistency within each cluster. Reasonable interpretations can be given. Cells are clustered according to the semantic features such as the chromatin openness, the darkness and density of nuclei, and if nuclei show the appearance of being segmented.

{When it comes to unsupervised classification, none of the baseline methods can benefit from the bounding box. We observe that the color context of the background can be disturbing when the classification is under the fully unsupervised manner.}

\begin{table}[H]
\centering
\caption{\textnormal{Performance of cell-level clustering.}}
\label{table:cluster}%
\resizebox{1\hsize}{!}{
\begin{tabular}{|c|c|l|c|c|c|c|c|c|}
\hline
\multirow{2}{*}{Dataset} & \multicolumn{2}{c|}{\multirow{2}{*}{Methods}} & \multicolumn{2}{c|}{Purity} & \multicolumn{2}{c|}{Entropy} & \multicolumn{2}{c|}{F-score} \\ \cline{4-9} 
                         & \multicolumn{2}{c|}{}                         & \tiny{{w/ Seg}}           & \tiny{{w/o Seg}}  & \tiny{{w/ Seg}}           & \tiny{{w/o Seg}}   & \tiny{{w/ Seg}}           & \tiny{{w/o Seg}}   \\ \hline
\multirow{4}{*}{A}       & \multicolumn{2}{c|}{MF+k-means}               & 0.579            &{0.442}     & 1.376            &{1.598}      & 0.603            &{0.510}      \\
                         & \multicolumn{2}{c|}{DNN+k-means}              & 0.667            &{0.470}     & 1.256            &{1.552}      & 0.677            &{0.501}      \\
                         & \multicolumn{2}{c|}{HOG+DEC}                  & 0.729            &{0.637}     & 1.086            &{1.167}      & 0.737            &{0.664}      \\
                         & \multicolumn{2}{c|}{Our Method}               & \textbf{0.855}   & {/}        & \textbf{0.750}   &{ /}         & \textbf{0.863}   & {/}         \\ \hline
\multirow{4}{*}{B}       & \multicolumn{2}{c|}{MF+k-means}               & 0.392            &{0.421}     & 1.561            &{1.545}      & 0.409            &{0.454}      \\
                         & \multicolumn{2}{c|}{DNN+k-means}              & 0.719            &{0.406}     & 0.844            &{1.557}      & 0.760            &{0.435}      \\
                         & \multicolumn{2}{c|}{HOG+DEC}                  & 0.771            & {0.681}       & 0.697            &{1.161}          & 0.812            &{0.693}          \\
                         & \multicolumn{2}{c|}{Our Method}               & \textbf{0.874}   & {/}        & \textbf{0.431}   & {/}         & \textbf{0.841}   & {/}         \\ \hline
\multirow{4}{*}{C}       & \multicolumn{2}{c|}{MF+k-means}               & 0.459            &{0.446}     & 1.533            &{1.597}      & 0.484            &{0.514}      \\
                         & \multicolumn{2}{c|}{DNN+k-means}              & 0.578            &{0.458}     & 1.377            &{1.575}      & 0.601            &{0.485}      \\
                         & \multicolumn{2}{c|}{HOG+DEC}                  & 0.667            & {0.602}     & 1.217            &{1.334}       & 0.682            &{0.621}      \\
                         & \multicolumn{2}{c|}{Our Method}               & \textbf{0.769}   & {/   }     & \textbf{0.977}   & {/   }      & \textbf{0.777}   &{ /   }      \\ \hline

\end{tabular}}
\end{table}

\begin{figure}[H]
  \subfigure[]{ 
    \label{loss} 
    \begin{minipage}{0.23\textwidth} 
      \includegraphics[width=\textwidth]{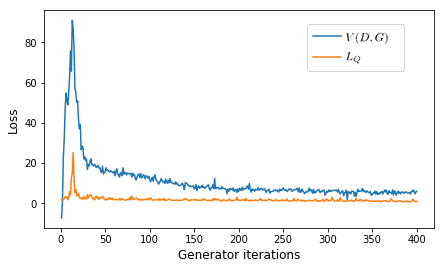}
    \end{minipage}}
  \subfigure[]{ 
    \label{purity}
    \begin{minipage}{0.23\textwidth} 
      \includegraphics[width=\textwidth]{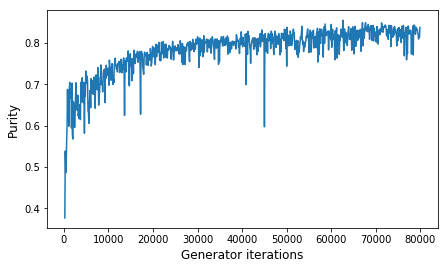}
    \end{minipage}} 
  \label{fig:mini:subfig}
  \caption{Visualization of cell-level clustering performed on Dataset A: (a) Training losses converge as the network trains. (b) The purity increases gradually over generator iterations.}
\end{figure}

Especially for Dataset A, Figure~\ref{loss} shows the convergence of $V(D,G)$ (see Equation (\ref{fig:wgan})) and $L_Q$ (see Equation (\ref{fig:InfoGAN})). $V(D,G)$ is used to evaluate how well the generator distribution matches the real data distribution \cite{gulrajani2017improved}. $L_Q$ approaching zero indicates that mutual information is maximized \cite{chen2016infoGAN}. Figure~\ref{purity} shows how the purity of clustering increases in the training process. 

\textbf{Impacts of the Number of Clusters:} 
For our method, it is easy to change the number of clusters by sampling the categorical noise from a different dimension. We compare the results of choosing different numbers of clusters shown in Table~\ref{different_k}, which shows there is no distinct difference between choosing four and five clusters. We choose five clusters (a 5-dimensional categorical random variable) in change for a slightly better performance.

\begin{table}[H]
\centering
\caption{\textnormal{Performance when choosing different numbers of clusters.}}
\label{different_k}
\begin{tabular}{|c|c|c|c|}
\hline
Clusters & 4     & 5     & 6     \\ \hline
F-score  & 0.831 & 0.863 & 0.789 \\ \hline
\end{tabular}
\end{table}

{{
\textbf{Impacts of Uninformative Representations}: The uninformative representations such as the staining color and rotations can be interference factors in the process of classification. Besides using color normalization and data augmentation to ease this problem, we also demonstrate that these features are more likely to be latent encoded in Gaussian random variables which do not influence the classification task. As is shown in Figure~\ref{fig:Uninformative}, we fix the value of the chosen categorical variable $c$ while walking through the random space of the Gaussian noise variable $z$. The result shows that uninformative representations tend to be encoded in noise variables through the process of maximizing the mutual information.

\begin{figure}[H]
\centering
  \subfigure[]{ 
    \label{staining}
    \begin{minipage}{0.23\textwidth} 
      \includegraphics[width=\textwidth]{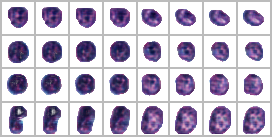} 
    \end{minipage}} 
  \subfigure[]{ 
    \label{rotation}
    \begin{minipage}{0.23\textwidth} 
      \includegraphics[width=\textwidth]{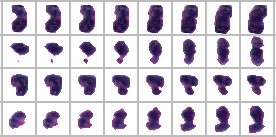} 
    \end{minipage}} 
  \caption{{{Examples of how uninformative representations are encoded in Gaussian noise variables $z$. Different columns share the same value of the chosen categorical variable $c$. A random walk is performed between two points in the space of $z$. It can be seen that (a) the staining color and (b) the rotation are both latent encoded in the Gaussian noise variables.}}}
  \label{fig:Uninformative} 
\end{figure}
}}

\subsection{Image-level Classification}

{{We perform image-level classification experiments on Dataset C and Dataset D respectively. Dataset C includes 29 positive and 11 negative images. Dataset D includes 132 positive and 72 negative images. Each dataset is randomly split into four folds for the 4-fold cross-validation. Each score is reported averagely. Each experiment is repeated for four times with different random split for cross-validation. The scores are reported four times to show confidence intervals.}}

{\textbf{Comparison:} (1) DNN (cell-level based): We use ResNet-50 features extracted from cell-level instances to perform cell-level clustering. Then we train an L2-SVM on top of the cell proportions to perform image-level classification. (2) DNN (image-level based): We use ResNet-50 pre-trained on Imagenet-1K as an image-level feature extractor. Images with a resolution of $1500 \times 800$ are normalized and center cropped to $800 \times 800$ pixels, then resized into $224 \times 224$ pixels. An L2-SVM is trained on the feature vectors. We observe this produces a better result than fine-tuning or directly training a ResNet-50 without pre-train.} (3) Our method (w/ k-means): We first train our GAN architecture on the training set, then conduct the cell-level clustering on both the training set and test set using the trained model. Cluster centers are calculated given cell proportions of each sample in the training set. The predict label is given by the closest cluster that each sample in the test set belongs to. (4) Our method (w/ SVM): An L2-SVM instead of k-means is used as the final classifier.

\textbf{Evaluation:} We use the precision, recall and F-score for evaluation, the details of which have been described in Equation \ref{eq:fscore}. The difference is that the labels are binary in this experiment.


{{\textbf{Results:} Following the proposed pipeline, the GAN architecture is trained on the segmentation output of the split training set. For cell-level clustering task, we achieve 0.791 F-score trained on 12000 training instances of Dataset C and 0.771 F-score trained on 60000 training instances of Dataset D, both evaluated by labeled cells of Dataset A. 

Given the cell proportions, when using k-means to perform image-level unsupervised classification, we achieve 0.931 F-score on Dataset C and 0.875 F-score on Dataset D, which is comparative to the DNN method with 0.933 and 0.888 F-score. The advantage is that our model is interpretable. The proportion of which category of cells is irregular is recognizable.

Since there are a large number of cell-level images on both Dataset C and D, it is difficult to test our method under full-supervision with a similar pipeline. We instead train an L2-SVM on cell proportions, taking image-level labels of histopathology images as targets. As the comparison shown in Table~\ref{imagelevel}, our method achieves 0.950 F-score on Dataset C and 0.902 F-score on Dataset D.}}

\begin{figure}[tbh]
\includegraphics[width=0.46\textwidth]{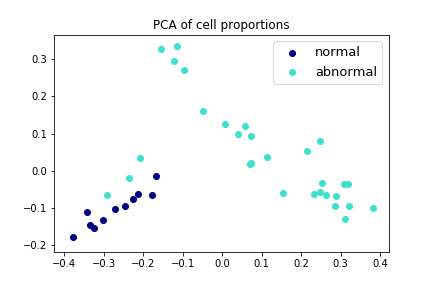} 
\caption{Visualization of unsupervised classification using cell proportions. It can be observed that the points representing normal and abnormal samples are distinctly distributed in two different clusters.} 
\label{visualization} 
\end{figure}

\begin{table*}[tbh!]
\centering
\caption{{\textnormal{Performance of image-level classification. Each experiment is repeated for four times with different random split for cross-validation. The scores are reported four times to show confidence intervals.}}}
\label{imagelevel}
\begin{tabular}{|c|c|c|l|l|l|c|l|l|l|c|l|l|l|}
\hline
Datasets           & Methods                   & \multicolumn{4}{c|}{Precision}         & \multicolumn{4}{c|}{Recall}            & \multicolumn{4}{c|}{F-score}           \\ \hline
\multirow{4}{*}{C} & {DNN (cell-level based)} &{0.539}&{0.598}&{0.688}&{0.524}      
                                       &{0.711}& {0.723}&{0.734}&{0.678}
                                       &{0.636}&{0.678}&{0.701}&{0.621}       \\
                   & DNN (image-level based)&0.906 & {0.913} & {0.901} &{0.921} 
                                      & \textbf{0.969} &{0.958} &{0.943} &{0.965} 
                                      & 0.933 & {0.929} & {0.924} & {0.937} \\
                   & Our Method (w/ k-means) & 0.936          & {0.945}  & {0.939} &{0.937} &0.933          &{0.944} &{0.946} &{0.938} &0.931          &{0.941}   &{0.948}  & {0.939} \\
                   & Our Method (w/ SVM)     & \textbf{0.950} & {\textbf{0.948}} &{\textbf{0.940}} & {\textbf{0.946}} &\textbf{0.969} &{\textbf{0.968}} & {\textbf{0.950}} &{\textbf{0.966}} & \textbf{0.950} &{\textbf{0.949}}  &{\textbf{0.940}}  &{\textbf{0.949}}    \\ \hline
\multirow{4}{*}{{D}} & {DNN (cell-level based)}  &{0.469}&{0.579}&{0.498}&{0.581}                                             &{0.697}&{0.654}&{0.643}&{0.665}                                             &{0.558}&{0.612}&{0.583}&{0.621}       \\
                   & {DNN (image-level based)}         &{0.863}&{\textbf{0.900}}&{0.887}&{0.869}   &{\textbf{0.863}}&{0.886}&{0.871}&{0.865}   &{\textbf{0.863}}&{0.888}&{0.879}&{0.866}  \\
                   & {Our Method (w/ k-means)} &{0.858}           &{0.879}  &{0.881}  &{0.868}   &{0.857}           &{0.868}  &{0.873}  &{0.865}   &{0.862}           &{0.870}  &{0.875}  &{0.867}  \\
                   & {Our Method (w/ SVM)}     &{\textbf{0.864}}           &{0.897}  &{\textbf{0.901}}  &{\textbf{0.882}}   &{0.858}           &{\textbf{0.892}}  &{\textbf{0.898}}  &{\textbf{0.878}}   &{\textbf{0.863}}           &{\textbf{0.891}}  &{\textbf{0.902}}  &{\textbf{0.880}}  \\ \hline
\end{tabular}
\end{table*}
On Dataset C, we use Principal Components Analysis (PCA) to perform a dimensionality reduction, cell proportions of each histopathology image are projected onto a two-dimension plane to show that there is a distinct difference between normal and abnormal images, shown in Figure~\ref{visualization}.

{{\textbf{Impacts of the Segmentation Parameters:} To validate the impacts of the segmentation performance on the image-level classification result, we change the value of intensity threshold in the segmentation process of experiments on Dataset C. We randomly choose 20 patches with a resolution of $200 \times 200$ pixels in Dataset C for evaluation, which includes 335 nuclei as counted. We use missing instances (nuclei that are missing in outputs), false alarms (mis-segmented background instances), and the F-score for evaluation. 

As is shown in Table~\ref{intensity}, both results of segmentation and classification are the highest when the intensity threshold remains 120. Followed by the decreasing of segmentation performance, the classification performance will stay within an acceptable range. Too bad segmentation performance will worsen the classification result since the quality and quantity of the segmentation outputs are not enough to reveal the distinct representation of each image-level instance.}}
\begin{table}[H]
\centering
\caption{\textnormal{{Performance when changing the segmentation parameters.}}}
\label{intensity}
\resizebox{1\hsize}{!}{
\begin{tabular}{|c|c|c|c|c|c|c|c|}
\hline
{\tiny{Intensity threshold}}                  & {60}    & {80}    & {100}   & {120}   & {140}   & {160}   & {180}   \\ \hline
{\tiny{Missing Instances}} & {127} & {48} & {21} & {7} & {14} & {64} & {184} \\ \hline
{\tiny{False Alarms}} & {3} & {4} & {15} & {5} & {20} & {30} & {35} \\ \hline
{\tiny{Segmentation F-score}}          & {0.315} & {0.413} & {0.602} & {0.701} & {0.656} & {0.534} & {0.218} \\ \hline
{\tiny{Classification F-score}} & {0.579} & {0.814} & {0.932} & {0.950} & {0.941} & {0.901} & {0.576} \\ \hline

\end{tabular}}
\end{table}

{{\textbf{Impacts of the Number of Clusters:} For image-level classification of Dataset C, we conduct experiments choosing different number of clusters. Table~\ref{image-level-cluster} shows that there is no distinct difference of performance between choosing five and six clusters. We still choose five clusters for a better performance.}}

\begin{table}[H]
\centering
\caption{\textnormal{{Performance when choosing different numbers of clusters.}} }
\label{image-level-cluster}
\begin{tabular}{|c|c|c|c|c|}
\hline
{\tiny{Clusters}}& {4}     & {5}     & {6}     & {7}     \\ \hline
{\tiny{Cell-level Classification F-score}}  & {0.711} & {0.791} & {0.762} & {0.710} \\ \hline
{\tiny{Image-level Classification F-score}} & {0.897} & {0.950} & {0.944} & {0.899} \\ \hline
 
\end{tabular}
\end{table}

{\textbf{Patch-level Classification}: We perform classification based on patches. Using a sliding window with a window size of 224 and a stride of 224, we separately transfer the normalized images from the training set and test set from Dataset C into labeled image patches. This results in 588 positive and 288 negative patches for training, 224 positive and 108 negative patches for testing. If 50\% of the patches of an image-level instance are positive, we will consider this instance as positive. In this manner, we achieve 0.851 F-score using DNN feature extractor with SVM and 0.831 F-score using our method, which is not comparative to our image-level classification results.}

\textbf{Discussion:}
Analyzing the results, we find that the cell proportions \{$P_1, P_2, \cdots, P_5$\} can indicate the presence of blood diseases.

For our experiment, cell-level clustering shows that \{$P_1$, $P_4$\} correspond to myeloblasts, \{$P_5$\} corresponds to lymphocytes and erythroid precursors, and \{$P_2$, $P_3$\} correspond to monocytes and glanulocytes. For all normal images, $P_1$ and $P_4$ are relatively lower. This matches the constitution in normal bone marrow where the lymphocytes, glanulocytes and erythroid precursors are in the majority when the percentage of cells with open phased nuclei (such as myeloblasts, under some circumstances plasma cells) is relatively lower (less than 10\%). In Figure~\ref{visualization}, abnormal images that are confidently discriminated are reflected in the numerous presence of the supposed minority myeloblasts or plasma cells, which in turn is reflected in the sharp increase of $P_1$ and $P_4$. 

\begin{figure}[H]
\centering
\includegraphics[width=0.3\textwidth]{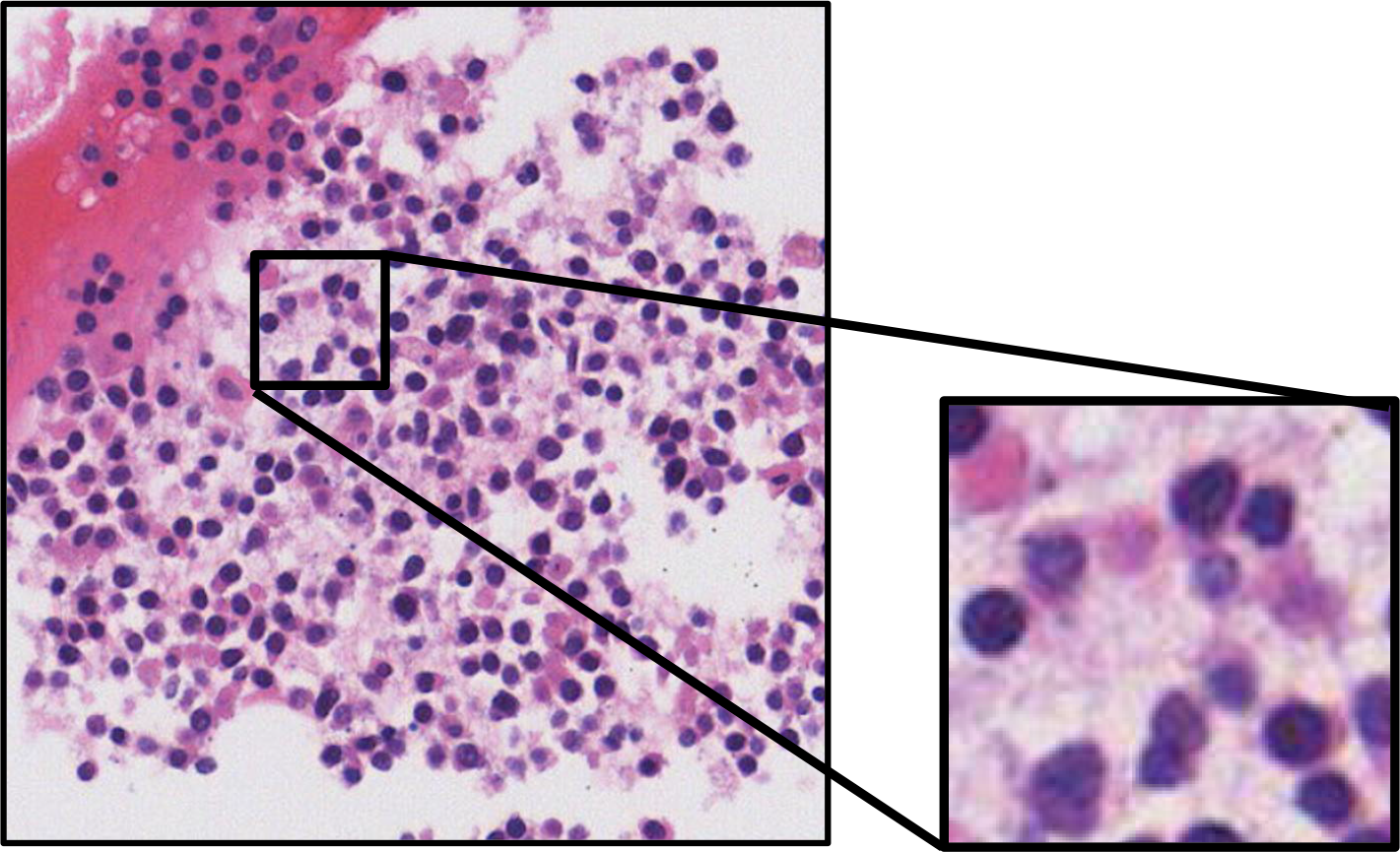} 
\caption{Example of the failed samples. Too many erythroid precursors indicate the presence of blood disease. The overlap of nuclei and the lousy staining condition add to the difficulties of cell-level classification.}
\label{wrong} 
\end{figure}

However, there are three abnormal images that are exceptional. To analyze what causes the failure, we display the example image in Figure~\ref{wrong}.

\begin{figure}[H]
\centering
  \subfigure[]{ 
    \label{dataset2_lymph}
    \begin{minipage}{0.23\textwidth} 
      \includegraphics[width=\textwidth]{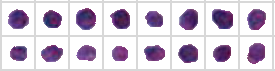} 
    \end{minipage}} 
  \subfigure[]{ 
    \label{dataset2_yuanshi}
    \begin{minipage}{0.23\textwidth} 
      \includegraphics[width=\textwidth]{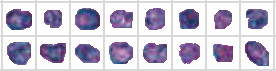} 
    \end{minipage}} 
  \subfigure[]{ 
    \label{dataset1_lymph}
    \begin{minipage}{0.23\textwidth} 
      \includegraphics[width=\textwidth]{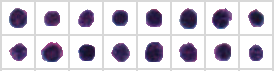} 
    \end{minipage}} 
  \subfigure[]{ 
    \label{dataset1_yuanshi}
    \begin{minipage}{0.23\textwidth} 
      \includegraphics[width=\textwidth]{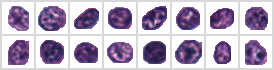} 
    \end{minipage}} 
  \caption{Variance of staining conditions. $(a)$ and $(b)$ are erythroid precursors and myeloblasts randomly chosen from failed images. $(c)$ and $(d)$ are samples selected from correctly predicted images. Our model mistakes erythroid precursors for myeloblasts particularly in failed images.} 
  \label{difference} 
\end{figure}

In these images, the irregular proportion of erythroid precursors indicates the presence of blood disease. We find that our model does not correctly classify these cells. The reason could be that the staining condition of these cells is not as good as expected. A typical erythroid precursor should have a close phased, dark-staining nucleus that appears almost black. As Figure~\ref{difference} shows, the color of nuclei segmented from these images differ from the rest of the dataset. Particularly in these images, our model is still not robust enough to capture the most significant semantic variance in an unsupervised setting. Therefore, acquiring high-quality histopathology images is still a priority.

\section{Conclusion} 

In this paper, we introduce a unified GAN architecture with a new formulation of the loss function into cell-level visual representation learning of histopathology images. Cell-level unsupervised classification with interpretable visualization is performed by maximizing mutual information. Based on this model, we exploit cell-level information by calculating the cell proportions of histopathology images. Followed by this, we propose a novel pipeline combining cell-level visual representation learning and nuclei segmentation to highlight the varieties of cellular elements, which achieves promising results when tested on bone marrow datasets.

In future work, some improvements can be made to our method. First, the segmentation method and the computational time can be further improved. The gradient penalty added on the network architecture requires the computation of the second order derivative, which is time-consuming in the training process. Secondly, in addition to cell proportions, other information about the patients should be carefully considered, such as clinical trials and gene expression data. By allocating and annotating the relevant genetic variants, the risk can be re-evaluated. In clinical practice, doctors need to consolidate more critical information to make a confident diagnosis. For example, bone marrow cells of children might not be as varied as those of adults'. To classify cells in a more fine-grained manner, the peculiar distribution information such as erythroid cells more likely form clusters (erythroid islands) can be considered. 

\section{Acknowledgment} 
The authors would like to thank the First Affiliated Hospital of Zhejiang University and Dr. Xiaodong Teng from Department of Pathology, the First Affiliated Hospital of Zhejiang University for providing data and help.
\medskip
\bibliographystyle{plain}

\end{document}